%% file: egpaper.tex
\documentclass[10pt,twocolumn,letterpaper]{article}

\usepackage{wacv}
\usepackage{times}
\usepackage{epsfig}
\usepackage{graphicx}
\usepackage{amsmath}
\usepackage{amssymb}
\usepackage[utf8]{inputenc} 
\usepackage[T1]{fontenc}    
\usepackage{url}            
\usepackage{booktabs}       
\usepackage{amsfonts}       
\usepackage{nicefrac}       
\usepackage{microtype}      

\usepackage{times}
\usepackage{epsfig}
\usepackage{graphicx}
\usepackage{amsmath}
\usepackage{amssymb}
\usepackage{adjustbox}
\usepackage{bm}
\usepackage{booktabs}
\usepackage{multirow}
\usepackage{multicol}
\usepackage{mathtools}
\usepackage{caption}
\usepackage{subcaption}
\usepackage[section]{placeins}
\usepackage[ruled]{algorithm2e}
\usepackage[dvipsnames]{xcolor}
\usepackage{array, caption, floatrow, tabularx, makecell, booktabs}%
\usepackage{pifont}
\newcommand{\xmark}{\ding{55}}%

\wacvfinalcopy 
\ifwacvfinal
\def\assignedStartPage{1} 
\fi

\ifwacvfinal
\usepackage[breaklinks=true,bookmarks=false]{hyperref}
\else
\usepackage[pagebackref=true,breaklinks=true,colorlinks,bookmarks=false]{hyperref}
\fi

\ifwacvfinal
\else
\fi

\begin{document}
\newcommand{\method}{TS-GAN~}
\newcommand{\methodshort}{NT}
\newcommand{\methodlong}{Temporal Shift GAN}

\title{Temporal Shift GAN for Large Scale Video Generation}

\author{Andres Munoz\thanks{Equal Contribution} , Mohammadreza Zolfaghari\footnotemark[1] , Max Argus, Thomas Brox\\
University of Freiburg\\
{\tt\small $\lbrace$amunoz, zolfagha, argusm, brox$\rbrace$}{\tt\small@informatik.uni-freiburg.de}}

\maketitle

\begin{abstract}
Video generation models have become increasingly popular in the last few years, however the standard 2D architectures used today lack natural spatio-temporal modelling capabilities. In this paper, we present a network architecture for video generation that models spatio-temporal consistency without resorting to costly 3D architectures. The architecture facilitates information exchange between neighboring time points, which improves the temporal consistency of both the high level structure as well as the low-level details of the generated frames. The approach achieves state-of-the-art quantitative performance, as measured by the inception score on the UCF-101 dataset as well as better qualitative results.
We also introduce a new quantitative measure (S3) that uses downstream tasks for evaluation. Moreover, we present a new multi-label dataset \textit{MaisToy}, which enables us to evaluate the generalization of the model.
\end{abstract}

\section{Introduction}
\vspace{-3mm}


\begin{figure}[h!]
\centering
\begin{subfigure}{.6\textwidth}
  \centering
  \includegraphics[width=.9\linewidth]{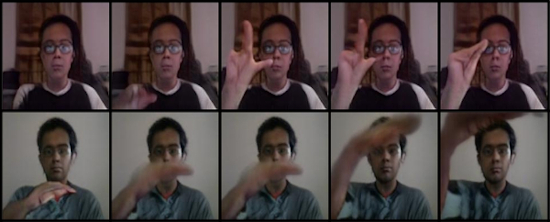}
  \caption{Jester Samples.}
  \label{fig:sub1}
\end{subfigure}%
\begin{subfigure}{.4\textwidth}
  \centering
  \includegraphics[width=.86\linewidth]{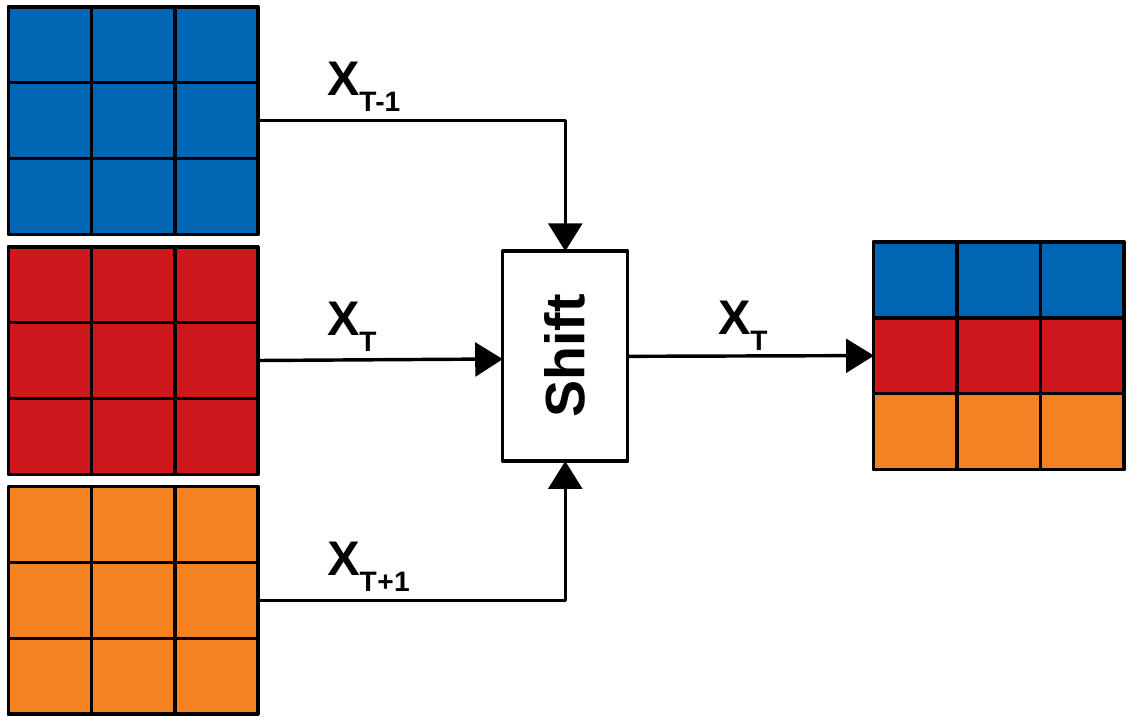}
  \caption{Temporal Shift.}
  \label{fig:sub2}
\end{subfigure}
\caption{(a) Selected frames from videos generated by TSB trained on Jester at 192$\times$192. (b) The shift operation replaces a subset of features in time step $T$ with features from frames $T-1$ and $T+1$ to facilitate information exchange between neighboring frames.}
\vspace{-.5em}
\label{fig:generator_blocks}
\end{figure}

\vspace{-.5em}
Generative Adversarial Networks (GANs) \cite{origigan} are a powerful way to generate high-resolution images~\cite{style, biggan}. 
Video generation adds further complexity, as the resulting content should be both spatially and temporally coherent. This is particularly true for the aspect of motion, which does not exist in still images.

3D Convolutional Neural Network (CNN) architectures appear well-suited to trivially lift the progress made in single images to videos~\cite{c3d, quo_vadis, eco, hara3dcnns}, yet their usefulness for video generation is still a matter of debate~\cite{moco, tgan}.
A argument against 3D CNNs is that the temporal dimension behaves differently from the spatial dimensions. The authors of MoCoGAN~\cite{moco} showed that equal treatment of space and time results in fixed-length videos, whereas the length of real-world videos varies. Moreover 3D CNNs have more parameters, which according to studies in  literature~\cite{tsm,brain_gan} make them more susceptible to overfitting~\cite{dm_h}.

We share the view of TGAN~\cite{tgan} and MoCoGAN~\cite{moco}, where instead of mapping a single point in the latent space to a video, a video is assumed to be a smooth sequence of points in a latent space in which each point corresponds to a single video frame. As a result, our video generator consists of two submodules: a \emph{sequence generator} that generates a sequence of points in the latent space, and an \emph{image generator} that maps these points into image space.

For the image generator, we propose a Temporal Shift Self-Attention Generator, which introduces a temporal shifting mechanism~\cite{tsm} into residual blocks of the generator. Temporal shifting mechanism enables the model to exchange information between neighbor frames. The temporal shifting module is complementary to 2D convolutions in the image generator and allows us to efficiently model the temporal dynamics of a video by facilitating the information exchange between neighbor frames. 

The growing interest in video generation methods gives rise to challenges in comparing the quality of generated samples. There are two types of approaches to evaluation: qualitative and quantitative. On one side, qualitative measures (e.g. human rating) are not good at detecting memorization or low diversity. On the other side, quantitative measures are not robust nor consistent~\cite{isbad, borji_eval_gan, lucic_17}. Although IS~\cite{inception} has gained popularity in evaluating the quality of generated images, it has several drawbacks; particularly failing to detect mode collapse and memorization. FID~\cite{frechet} assumes that features are from Gaussian distribution, which is not always a valid assumption. 

Therefore, we propose a new evaluation measure named Symmetric-Similarity-Score (S3) to measure the quality of generated videos. S3 measures the domain gap of an action classifier when trained on synthesized videos and tested on real ones, and vice-versa. Consequently, it penalizes missing intra-class diversity, and is also sensitive to both structural deficits and low-level artifacts in the generated data. Hence, it is robust to over-confident classifier predictions, and it is less dependent on model parameters or pre-processing. 

Currently video generation models have relied on action recognition datasets for benchmarking. However, as these datasets typically only assign one label per video, they do not allow for an easy analysis of the generalization capabilities of a model. By formulating the conditional generative modelling problem as a multi-label one, we can easily analyze generalization by forcing the model to try to generate samples from label combinations that are not in the dataset.

Experiments on the UCF101, Jester, and Weizmann datasets show substantial improvements in the quality of the generated videos for the proposed design compared to previous work. At the same time, experiments on the newly introduced MaisToy dataset show that \method is able to generalize to unseen data.



Our paper makes three contributions: (1) it introduces a new 2D video generator design with an ability to model spatio-temporal content by facilitating the information exchange between neighboring frames.
(2) It introduces a new evaluation metric based on the domain gap between synthesized and real videos in terms of video classification performance. (3) It introduces a new dataset which allows a fast and more in-depth analysis of the generalization and semantic modelling capabilities of video generation models.
\vspace{-4mm}

\section{Related Work}
\vspace{-2mm}
\textbf{Image generation} has recently seen leaps in performance \cite{infogan, progan, sngan, style, attention, biggan, deli}, thanks to recently introduced frameworks such as SN-GAN (Miyato \textit{etal} \cite{sngan}), introduced the concept of spectral normalization of the discriminator's weights. Zhang \textit{et al} \cite{attention}, designed a self-attention module that allowed the network to create non-local spatial relationships. Then, BigGAN \cite{biggan} build upon this work by establishing some architectural and training guidelines by which GANs can be stable, converge faster and produce better quality samples.  In this study we propose \method, which builds upon BigGAN and extends it to video.  \method generates videos in a per-frame basis, it can thus exploit further developments on image generation.

\textbf{Video generation} is a highly challenging task as a result of needing to ensure a smooth transition across video frames. 
Most  works in video generation have been on the closely related frame prediction task \cite{future1, future2, future3, future4, future5, future6, future7, future8}. The main difference between video generation and frame prediction, is that in frame prediction the network is trying to generate a set of $T$ frames given a set of $N$ previously seen frames. Conversely, video generation only uses the latent code, and in some occasions a label, to generate a set of frames. 

Several frameworks for video generation using GANs have been proposed in the past. Vondrick \textit{et al} \cite{vgan} proposed a two-stream network which explicitly separated the generation of the foreground and the background. They assumed that background in the entire video is static, which is not true in real-world video datasets. Saito \textit{etal} \cite{tgan} introduced a temporal generator to transform a single latent variable into a sequence of latent variables, to be able to just utilize a 2D network as a generator. As a matter of fact, they showed that a 2D generator can outperform a 3D one.
MoCoGAN \cite{moco} separated motion and appearance features by dividing the latent code in two smaller sub-codes, one per-each set of features. 

Acharya \textit{etal} \cite{progrevgan} used a coarse to fine approach to improve quality and convergence speed. TGANv2 \cite{tganv2}, efficiently trains models that generate high dimensional samples by subsampling features and videos from the batch. DVD-GAN \cite{dvdgan}, leveraged a high capacity model to synthesize high quality samples from complex datasets. These models showed that GANs can be effective at generating videos. Nevertheless, the previously proposed models either suffer from lack of quality, mode collapse, memorization or require an excessive amount of computational power and data to train properly. Our framework outperforms the state-of-the art on UCF-101, based on IS measure, while keeping memorization to a minimum.

Moreover, previous methods lack a complete quantitative assessment of the performance of the respective methods. They rely on metrics such as IS~\cite{is_score} and FID~\cite{fid_score} which don't tell the full story about sample quality. These metrics are dependent on availability of models and are also sensitive to changes in the pipeline. Here we introduce a metric, called Symmetric Similarity Score (S3), which aims to represent both quality and diversity in a single scalar value. In addition, S3 is robust to changes in pre-processing and model parameters. 
\vspace{-4mm}

\section{\methodlong}
\vspace{-2mm}
\subsection{Preliminaries}
\vspace{-1mm}
GANs \cite{origigan} are a class of generative models consisting of a generator and a discriminator networks. The discriminator is a binary classifier that outputs the probability a sample is either real or synthesized. The generator is a function that generates synthetic samples $x$ that look similar to real samples.

GAN training is a minimax game, in which the discriminator $D$ tries to minimize the probability of making a mistake, while the generator $G$ seeks to maximize this probability:
\vspace{-.4em}

\begin{equation}
    \begin{split}
        \mathop{\text{min}}_{G}\: \mathop{\text{max}}_{D}\: V(D,G) =\mathop{\mathbb{E}_{x \sim p_{data}(x)}}[log\:D(x)]\:\: +\\ \mathop{\mathbb{E}_{z \sim p_{z}(z)}}[log\:(1 - D(G(z)))]
    \end{split}
    \label{eq:gan}
\end{equation}
where $p_{data}$ is the distribution of the empirical data and $p_z$ represents the chosen prior distribution of the latent codes $z$.

Although GANs tend to have problems generating diverse samples (mode collapse), the recent BigGAN method~\cite{biggan} demonstrated state-of-the-art performance in image synthesis by leveraging the best practice of previous methods, such as spectral normalization and projection.

\begin{figure}[t]
\begin{center}
   \vspace{-.5em}
   \includegraphics[scale=0.34]{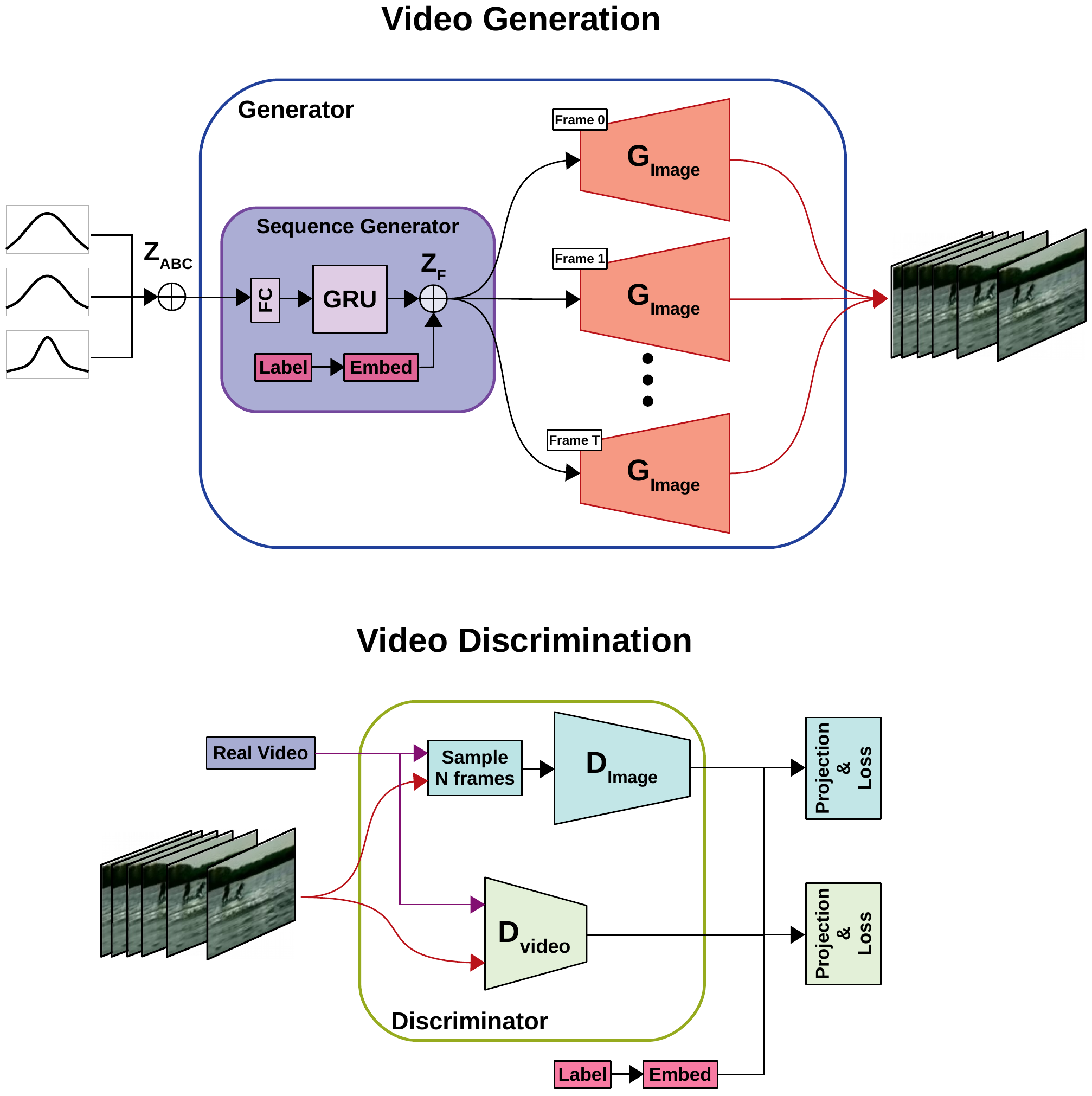}
\end{center}
   \caption{\method \: framework. 
   Sampled Gaussian noise is with several different variances is concatenated ($\oplus$) into $Z_{ABC}$. A sequence generator using gated recurrent units is then used to generate a vector $Z_F$. The image generator then transforms this into video frames. Video discrimination is done by a 2D discriminator ($D_\text{{Image}}$) judging a subset of the video frames and a 3D discriminator ($D_\text{{Video}}$) judging all frames and assessing the motion consistency of the video. }
   \vspace{-.8em}

\label{fig:arch}
\end{figure}

The proposed video generation architecture \method consists of a sequence generator, an image generator and a video discriminator; an overview of which is shown in Figure~\ref{fig:arch}. 
It is a projection based conditional GAN approach as proposed by Miyato \& Koyama \cite{cgans} using the hinge formulation of the GAN objective (Lim \& Ye \cite{geometric}; Tran et al.\cite{hierar}):
\vspace{-1mm}
\begin{equation}
    \begin{split}
         L_{D} = \mathop{\mathbb{E}_{(x,y) \sim p_{data}}} [\text{min}(0, -1 + D(x,y))] - \\ \mathop{\mathbb{E}_{z \sim p_{z}, y \sim p_{data}}} [\text{min}(0, -1 - D(G(z), y))]
    \end{split}
    \label{eq:hinge}
\end{equation}

\begin{equation}
    L_{G} =  - \mathop{\mathbb{E}_{z \sim p_{z}, y \sim p_{data}}} [D(G(z), y)]
    \label{eq:hinge2}
\end{equation}
where $y$ is the video label.
We introduce several improvements to different aspects of the video generating framework including sequence generator and the image generator.

\subsection{Generator}
\vspace{-1mm}
The generator is divided into two parts. First we generate a sequence of latent codes, then in the second step the image generator maps these latent codes to a sequence of $T$ frames.


\vspace{-2mm}
\subsubsection{Sequence Generator}
\vspace{-1mm}
We construct the latent space $Z_{ABC} \in \mathop{\mathbb{R}}^{d}$ as three independent multi-variate Gaussian distributions $Z_A \in \mathop{\mathbb{R}}^{d_A}$, $Z_B \in \mathop{\mathbb{R}}^{d_B}$ and $Z_C \in \mathop{\mathbb{R}}^{d_C}$ with their diagonal covariance matrices $\Sigma_A$, $\Sigma_B$ and $\Sigma_C$ respectively.
We construct our latent code, $Z_{ABC}$\footnote{This paper uses $Z_{ABC}$ to refer to the latent space, a vector of latent codes or a single latent code.}, by concatenation of ${Z_A,Z_B,Z_C}$ as $Z_{ABC} = \begin{bmatrix} Z_A, Z_B, Z_C \end{bmatrix}^T$.
The final distribution $Z_{ABC}$ is a multi-variate Gaussian distribution with diagonal covariance matrix. 
By using an independent parametrization of the subspaces, the network is able to learn more nuanced distributions, thus a better modelling of the features. Subspaces have no prior meaning - the network learns to interpret each part of the code as it sees fit.

The latent code $Z_{ABC}$ does not have a temporal dimension. Since our generator is image based, we first have to create a progression of correlated latent codes that extends through the intended duration of the video. 
This is done by the sequence generator (See Fig.~\ref{fig:arch}). We first transform the latent code with a fully connected layer as $Z_{fc} = FC(Z_{ABC})$.

Then we feed $Z_{fc}$ into a Gated Recurrent Unit (GRU) to generate a sequence of $T$ correlated codes as $Z_{gru} = [z^1_{gru},\ldots,z^T_{gru}]^T$, where each $z^i_{gru}$ corresponds to the $i$-th frame in the video sequence. In total this results in an input of size $[T, d]$, where $T$ is the number of frames to generate.

We concatenate these latent codes with per-class embeddings $e(y)$ of size $120$, where $y$ is a randomly sampled class label. This results in a sequence of $T$ codes as  
\vspace{-1mm}
\begin{equation}
    Z_F = \left [ \begin{bmatrix}
z^1_{gru}\\ e(y) \end{bmatrix}, \dots, \begin{bmatrix}
z^T_{gru}\\ 
e(y)
\end{bmatrix} \right ] \in \mathop{\mathbb{R}^{(d+120)}}
\end{equation}

We feed $Z_F$ into the image generator to generate a sequence of $T$ frames (Figure~\ref{fig:arch}). 
\vspace{-2mm}
\subsubsection{TSB Image Generator}
\begin{figure}[h!]
\centering
\begin{subfigure}{.85\textwidth}
  \centering
  \includegraphics[width=.9\linewidth]{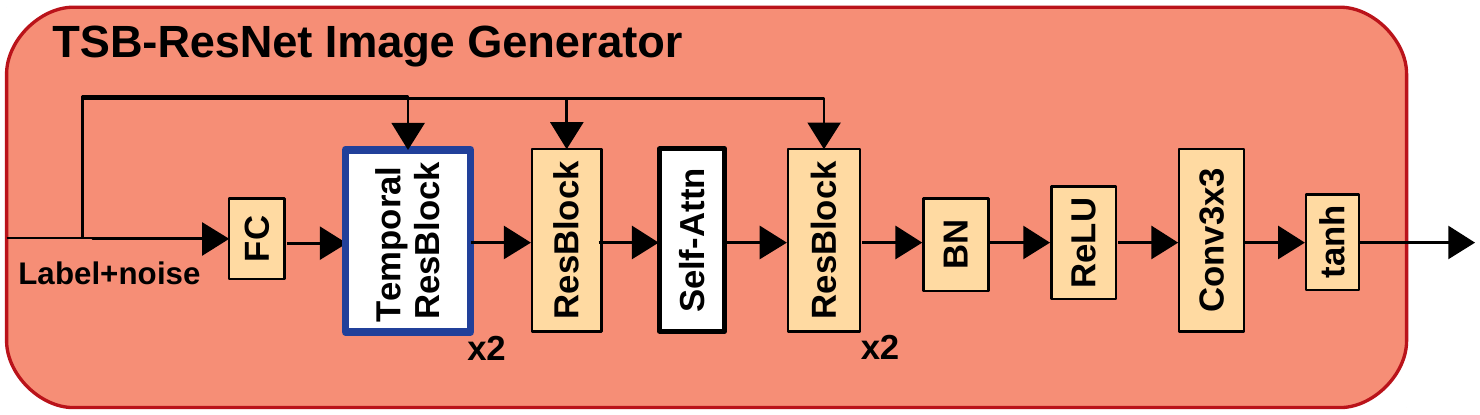}
  \caption{TSB Generator.}
  \label{fig:gen}
\end{subfigure}%
\vfill
\begin{subfigure}{.85\textwidth}
  \centering
  \includegraphics[width=.86\linewidth]{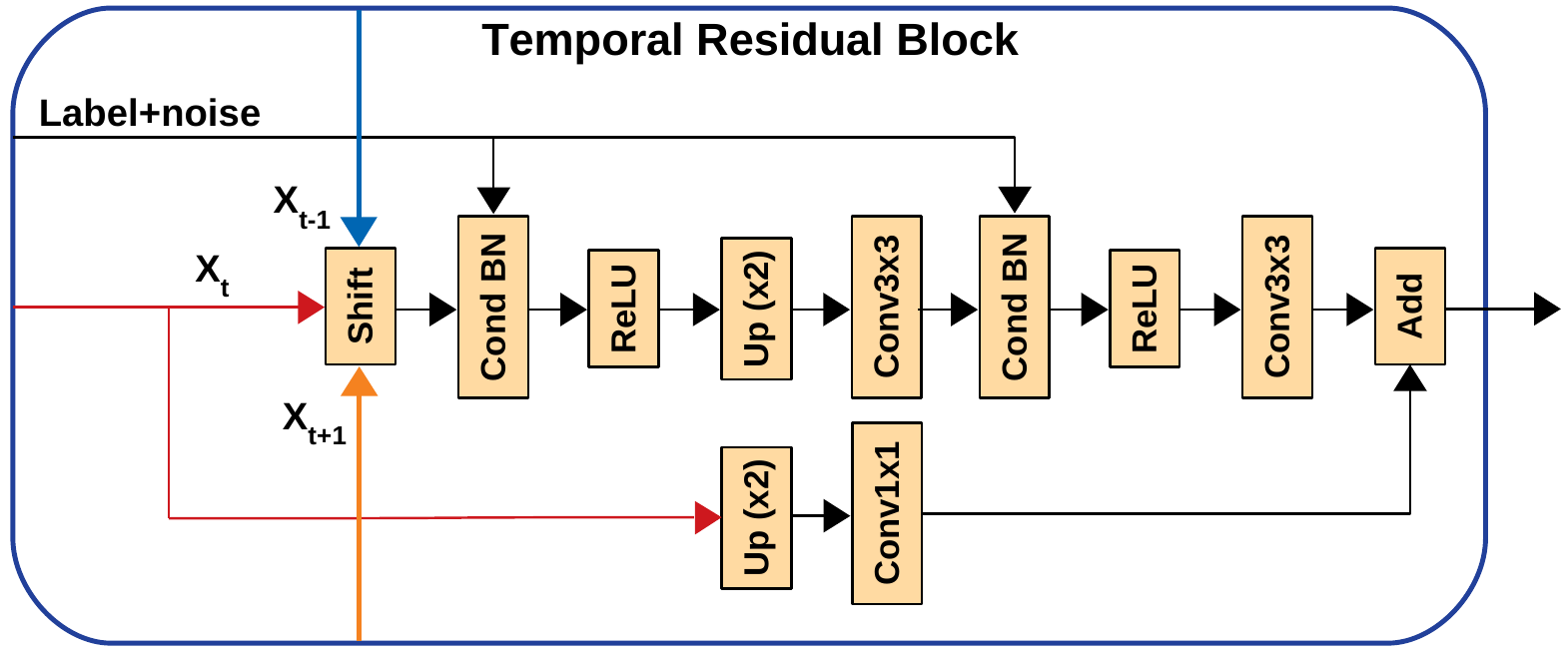}
  \caption{Temporal Residual Block.}
  \label{fig:block}
\end{subfigure}
\caption{(a) The Temporally Shifted (TSB) image generator architecture. Note that the temporal residual blocks are only used at the beginning of the generator to minimize loss of spatial information. (b) Temporal residual up-sampling block used in TSB. The operation Up(x2) means up sampling via interpolation by a factor of 2.}
\vspace{-.5em}
\label{fig:generator_blocks}
\end{figure}

\vspace{-1mm}
To synthesize ``realistic`` images, some approaches~\cite{dvdgan,jeff19gan} utilized BigGAN~\cite{biggan} image generator as their backbone architecture. However, in this architecture each image is generated independent of others. Therefore, the networks are not able to enforce temporal consistency between frames. To alleviate this problem, we introduce the temporal shift mechanism~\cite{tsm} to BigGAN image generator architecture to facilitate information exchange between neighboring frames. We call the proposed generator Temporally Shifted BigGAN (TSB) image generator, illustrated in Figure \ref{fig:gen}, because of it's feature shifting mechanism (Figure \ref{fig:sub2}). This design not only facilitates the information exchange in temporal dimension but also equipped with a self-attention layer which enables the generator to model the relationships between spatial regions~\cite{attention}.
Unlike full 3D convolutions it only shares a small subset of features between neighboring frames. This allows faster inference and uses less parameters than 3D models.


In our proposed image generator TSB, the temporal shift module can simply be added to the beginning of the residual blocks, as shown in Figure~\ref{fig:block}. This is in contrast to the non-temporal (\methodshort)~variant of our architecture, which uses normal residual blocks.
We only vary the first two residual blocks of our network, and call these \emph{temporal residual blocks} to distinguish them from the latter residual blocks which are always of the normal variant. This is shown in Figure~\ref{fig:gen}. All residual blocks use conditional batch normalization and receive as input the vector $Z_F$.

\vspace{-1mm}
\subsection{Discriminator}
\vspace{-1mm}
We use two independent discriminators, an image discriminator, $D_{\text{Image}}$, and a video discriminator named $D_{\text{Video}}$.

\textbf{Image Discriminator} $D_{\text{Image}}$ gives a frame-wise assessment of content and structure.
$D_{\text{Image}}$ is a ResNet based architecture~\cite{resnet}, similar to BigGAN \cite{biggan}, 
it is applied to a subset of $N$ frames of the video.
$D_{\text{Image}}$ is doing the heavy lifting with respect to image quality. $N$ remains a hyperparamter that allows a trade-off between memory efficiency and frame quality.

\textbf{Video Discriminator} $D_{\text{Video}}$ examines the temporal consistency of videos and provides the generator with a learning signal to generate a consistent motion throughout all $T$ frames. \method's $D_{\text{Video}}$ is inspired by MoCoGAN's \cite{moco} video discriminator. We chose this architecture to keep the network efficient. The factorized design allows for smaller $D_{\text{Video}}$ networks as it can focus on the temporal aspect.

\vspace{-1mm}
\section{Symmetric Similarity Score (S3)}
\label{sec:s3}
\vspace{-2mm}
The Inception Score \cite{inception} (IS) and Frechet Inception Distance \cite{frechet} (FID) are the most common metrics used to evaluate GANs. On one hand, IS ($\text{exp}(D_{KL}(P(y \: | \: x) \: | \: P(y)))$) is based on two criteria: the distribution of predicted labels $P(y \: | \: x)$ should have a low entropy and the marginal distribution $P(y)$ should have a high entropy. On the other hand, FID measures performance of a generator by using features produced by an intermediate layer to parameterize a multivariate normal distribution of real and fake features respectively. FID rates the fake samples by calculating the distance between distributions, the closer the better.

Although high IS correlates with subjective quality and a low FID with both quality and intra-class diversity of samples, they both have drawbacks. IS cannot capture intra-class diversity. Yushchenko \textit{etal} \cite{markov} showed that small changes to the data pre-processing leads to a change between 7\% and 17\% in IS score and adversarial samples may lead the classifier to be overconfident about samples leading to a higher score \cite{adversarial}. 
FID assumes features follow a normal distribution, which is not true for real world datasets. Thus, two completely different distributions might lead to a good score, while not being actually similar. At the same time, FID is also vulnerable to pre-processing and model changes. Neither IS nor FID are able to account for memorization of the dataset.
 
\textbf{Symmetric-Similarity-Score (S3)} uses generalization of  classifiers between real and synthetic samples to measure quality of generated videos. The performance of model is measured by "quality of samples" and "diversity of generated samples". 

\begin{figure*}
\begin{center}
\includegraphics[width=0.95\linewidth]{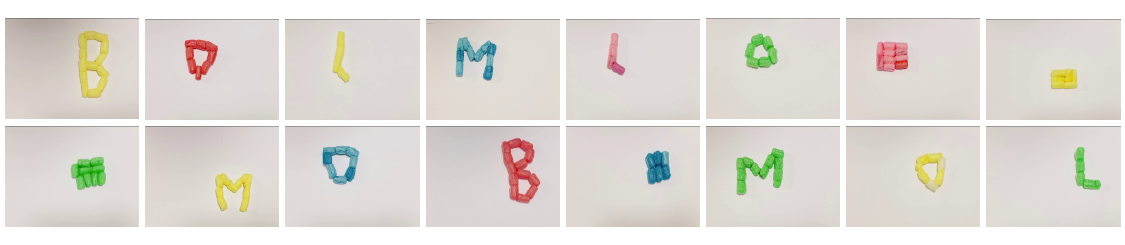}
\end{center}
\vspace{-.4em}
   \caption{Samples taken from MaisToy dataset. The dataset includes shapes of varying difficulty, the dataset is balanced in terms of colors and motions.}
   \vspace{-.8em}
\label{fig:maistoy}
\end{figure*}

The performance of a classifier trained on synthetic data and evaluated on real data (SeR) should increase, if synthetic samples are \textbf{diverse} and \textbf{realistic}. A classifier trained on real data being evaluated on synthetic (ReS) data should only perform well, if synthetic samples are \textbf{realistic}. 

We normalize these values by comparing to the real performance (ReR). Since SeR has more information about the overall performance, S3 has an exponential relationship to it, thus rewarding models with good diversity and sample quality and harshly penalizing them otherwise (Equation \ref{eq:s3}). S3 has the advantages of capturing intra-class diversity, being more robust to over-confident predictions and small changes in the model's parameters, while still being easier to interpret than IS or FID.

\vspace{.5em}
\begin{equation}
    S3 = \sqrt{  \left( \frac{SeR}{ReR} \right)^{2} \cdot \left(\frac{ReS}{ReR}\right) }
    \label{eq:s3}
\end{equation}

This approach is similar to Classification Accuracy Score (CAS)~\cite{class}, which used a classifier's SeR to evaluate generative models and lacks ReS evaluation. However, just using SeR to evaluate a model does not tell the full story. Since SeR is dependent on both quality of samples and intra-class diversity, we need ReS to know if the SeR performance is being driven by sample quality or diversity.

Generative models must create fake samples that comes from the same distribution as the dataset. To generate samples which are not included in the dataset it needs to be able to generalize to unseen data. However, existing datasets used for video generation are not truly equipped for testing a model's generalization due to the fact that we can only directly control the action semantic on the dataset. By only having control over the action, we can't force the network to generate certain features within the class, therefore it is hard to corroborate the model's ability to generalize. Thus, there's a need for a dataset that allows a higher degree of control over the semantics of the samples.
\vspace{-2mm}
\section{MaisToy Dataset}
\vspace{-2mm}
We introduce MaisToy, a dataset composed of 238 videos of clay figures of 5 different shapes, 4 colors performing 4 distinct motions. The videos recorded have 55 frames on average, with a size of $320 \times 240$ px.
The dataset is balanced and compact. This allows for faster evaluation of design choices without requiring large computational resources, this addresses a big challenge in video GAN research. The balanced nature of the dataset facilitates testing of generalization by holding out some combinations of semantics during training and trying to generate them during testing. At the same time, the three distinct semantics (shape, color and motion) support a more in-depth analysis of the semantic modelling capabilities of model designs.
\vspace{-2mm}
\section{Experiments}
\vspace{-2mm}
We evaluated our model both qualitatively and quantitatively on the quality of frames, the realism of whole videos, diversity and memorization using several different datasets. FID and further qualitative evaluation will be found in the supplementary material.

\subsection{Datasets}
\vspace{-2mm}
We use four datasets UCF101\cite{ucf101}, Weizmann\cite{weiz}, Jester\cite{jester} and MaisToy for our experiments.

\textbf{UCF-101.} $13,220$ videos of $101$ different sports action classes. We trained models to generate samples both at $96 \times 96$ and at $128 \times 128$, we resize to $127 \times 127$ and $170 \times 170$ respectively and crop to its corresponding final size. We set $N$ to 8. \cite{ucf101}

\textbf{Weizmann.} $93$ videos of $9$ people performing $10$ different actions. To train we cropped the videos to $96 \times 96$. For all experiments, we randomly extract 16 subsequent frames and set $N$ to 4. \cite{weiz}

\textbf{Jester.} $118,562$ videos and $27$ different hand gesture classes. Due to the small frame size, we first re-size $96 \times 136$ before cropping to $96 \times 96$ to preserve the aspect ration. As in Weizmann, we extract 16 subsequent frames and set $N$ to 4. \cite{jester}

$\textbf{MaisToy}_{\text{Multi}}$. Multi label variant of MaisToy, we trained a model to generate at $128 \times 128$, we resize as for UCF-101. We set $N$ to 4. For generalization testing, the dataset was split into train and test sets.

$\textbf{MaisToy}_{\text{Single}}$. Single label variant of MaisToy, we use only the motion labels to generate videos. We trained a model to generate at $96 \times 96$, we resize as for UCF-101. We set $N$ to 4.
\subsection{Model Configurations}
\vspace{-1mm}
Three different variation of our method were tested in order to find the strongest configuration.

\textbf{\methodshort:} A non-temporal model using a BigGAN generator and all latent variables' distribution were  $\mathcal{N}(\mu=0,\sigma=1)$.

\textbf{\methodshort-VAR:} Same generator as above, but we change the latent variables' distribution deviations to $\sigma_{A} = 0.5$, $\sigma_{B} = 1$ and $\sigma_{C} = 2$.

\input{table1}

\textbf{TSB:} Same as in \methodshort-VAR, however we change the generator to the temporally shifted BigGAN generator.

For all trained models we set $d$ to $120$ by setting $d_A$, $d_B$ and $d_C$ to 20, 20 and 80 respectively. We employ a  learning rate of $5\times10^{-5}$ for the generator and  $2\times10^{-4}$ for the discriminator. The video length $T$ is fixed to $16$.
All experiments performed on Weizmann and Jester are done with a batch size of 40.  We trained both \methodshort \: and TSB to generate $96 \times 96$ and $128 \times 128$ sized samples respectively. \methodshort \: was trained on a batch size of 56, while TSB was trained on a batch size of 64. 

All models were trained on full precision, $96 \times 96$ models were trained on two Nvidia RTX2080Ti's, $128 \times 128$ models were trained on one 32GB Nvidia V100. Jester and UCF-101 models took 4 weeks to reach the performance reported here, while models trained on Weizmann and MaisToy took 7 and 10 days respectively. Training times for Jester and UCF-101's could be cut short by using larger computational resources.

\begin{figure*}[h!]
\begin{center}
   \includegraphics[width=0.95\linewidth]{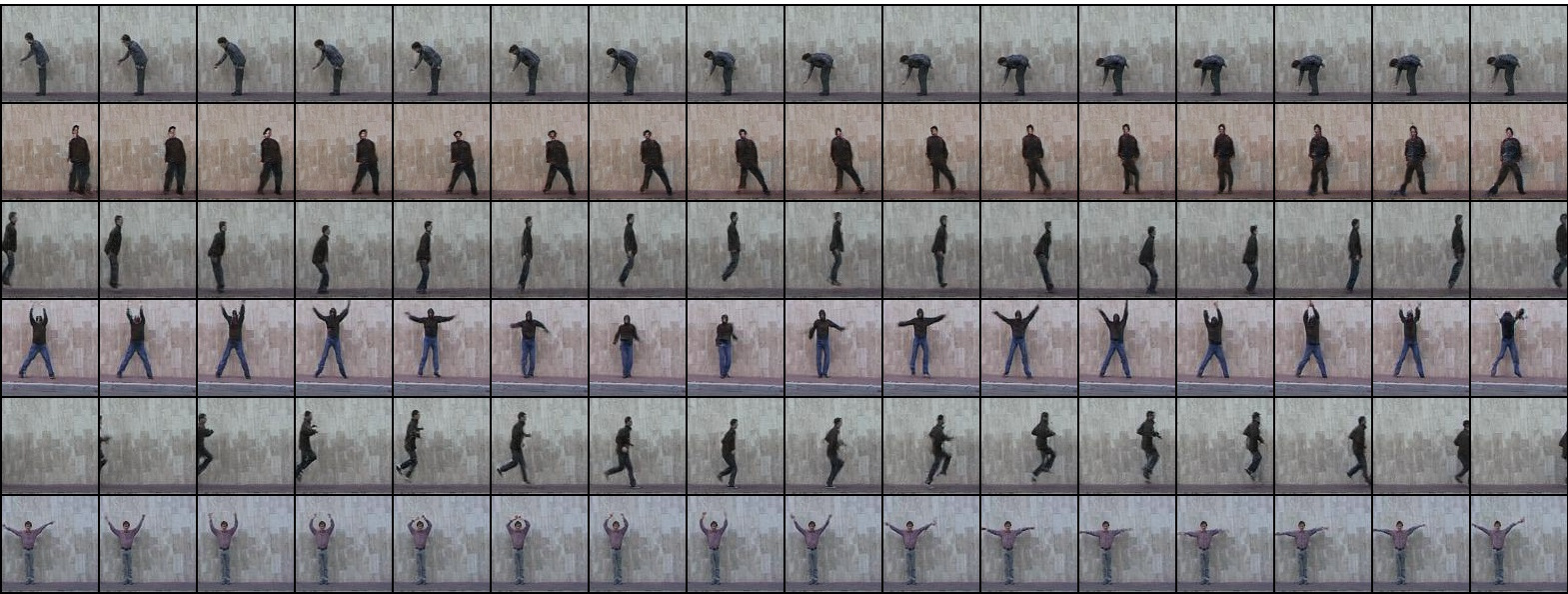}
\end{center}
   \caption{Generated samples on Weizmann dataset with our TSB model.}
   \vspace{-1em}
\label{fig:weiz_gen}
\end{figure*}

\subsection{Quantitative Evaluation}
\vspace{-1mm}
A thorough evaluation of quality of samples simply by qualitative experiments is not possible due to the sheer number of samples that need to be evaluated in order to do a proper assessment.

\begin{figure*}
\begin{center}
\includegraphics[width=0.95\linewidth]{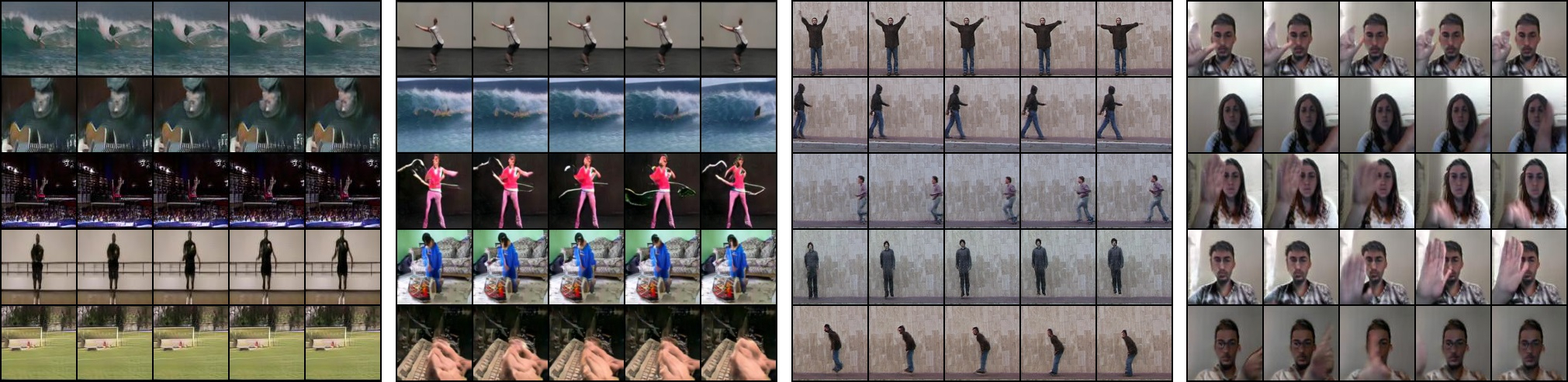}
\end{center}
\vspace{-.4em}
   \caption{Generated samples (from left to right) of \methodshort \: trained on UCF-101 and TSB trained on UCF-101, Weizmann and Jester. Videos can be found here: \url{https://drive.google.com/file/d/14k17fQTTztV2MPKAglOS6kP_yDKJ24n6/view?usp=sharing}}
   \vspace{-.8em}11
\label{fig:quality}
\end{figure*}

\textbf{IS:} We evaluate the IS as comparative benchmark on the UCF-101 dataset. The IS is calculated using the last layer of a C3D\footnote{Using the code provided by github.com/pfnet-research/tgan.} \cite{c3d} model which was pre-trained on Sports-1M \cite{sports} and fine-tuned on the UCF-101 dataset as per \cite{tgan}. The model receives frames of size $128\times128$, we resized when necessary. We use 10,000 samples to calculate the score. The standard deviation is calculated by repeating this procedure 10 times. Values for VGAN, MoCoGAN, Progressive VGAN, TGANv2, and DVD-GAN are shown as reported in \cite{dvdgan}, TGAN and MDP's values are reported as they appear in the original works \cite{tgan, markov}. On Jester, we use a TSN \cite{tsnet} action recognition network pre-trained on ImageNet \cite{imagenet} and fine-tuned on Jester, otherwise te same procedure as for UCF-101 is used. TSB produces samples that beat the state of the art (see Table \ref{tab:isscore}). Although, IS scores might suggest an overwhelming improvement over all existing methods, when qualitatively comparing samples from \methodshort~and TSB (Figure \ref{fig:quality}) we don't see a vast improvement as the score suggests. This could be because our samples might be exploiting C3D \cite{c3d} in a way that it is over confident about its prediction, thus a higher score.

\textbf{S3:} To calculate S3 on Weizmann and UCF-101 we used the TSN \cite{tsnet} action recognition network pretrained on ImageNet \cite{imagenet}. Since Jester is a motion dataset we decided to use ECO \cite{eco} because it incorporates a 3D network, to improve classification. On the Weizmann dataset we compare to MoCoGAN. All experiments on a dataset were done under the same conditions. Training details of the classifier will be included in the supplemental material. 
From Table~\ref{tab:classification_test}, we can see TSB produces a significant performance increase over all methods. It appears TSB, is able to increase the quality of the samples with a minimal loss of diversity. TSB was able synthesize test set samples of the $\text{MaisToy}_{\text{Multi}}$ dataset as implied the by SeR score being higher than the ReS, this suggests generalization. MaisToy samples will be included in the supplementary material. On UCF-101, Table~\ref{tab:classification_test} shows small  discrepancies between SeR and ReS indicating a good diversity of samples.
S3 scores produced by TSB show an improvement over \methodshort. The performance difference between methods seems more reasonable, when visually comparing samples (Figure \ref{fig:quality}), than the ones shown in Table \ref{tab:isscore}. Additionally, Table \ref{tab:s3_test} shows that S3 is able to to capture mode collapse in the generated samples, while still being equally as good as IS at measuring sample quality. This indicates that S3 is more reliable than IS. 

\subsection{Qualitative Results}
\vspace{-1mm}
\begin{figure}[h]
\begin{center}
   \includegraphics[width=0.95\linewidth]{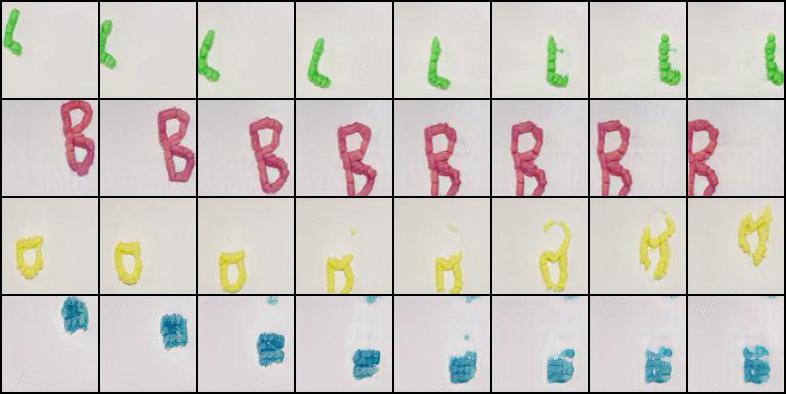}
\end{center}
   \caption{Samples from the $\text{MaisToy}_{\text{Single}}$ dataset, here we present samples from classes (top to bottom) right, left, up and down. The shapes represented here are Letter L, Letter B, triangle and square.}
   \vspace{-1em}
\label{fig:mais_normal}
\end{figure}

\begin{figure}
\centering
\begin{subfigure}{.95\textwidth}
  \centering
  \includegraphics[width=.95\linewidth]{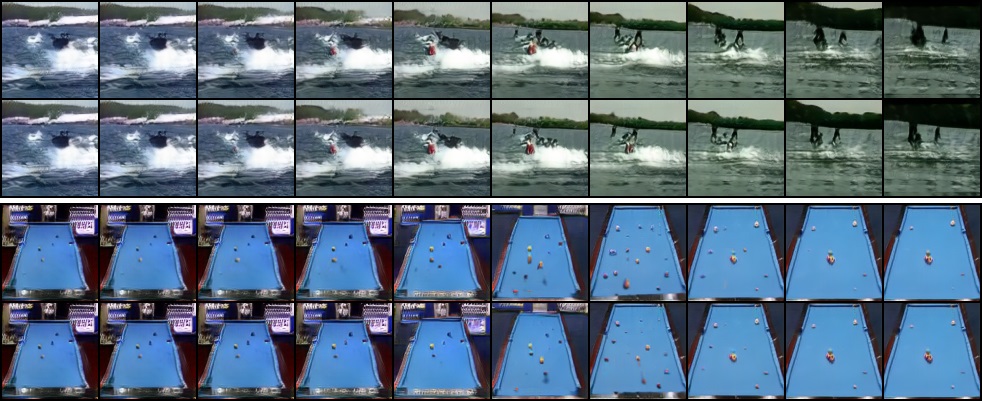}
  \caption{\methodshort.}
  \label{fig:tmm-intra1}
\end{subfigure}%
\vfill
\begin{subfigure}{.95\textwidth}
  \centering
  \includegraphics[width=.95\linewidth]{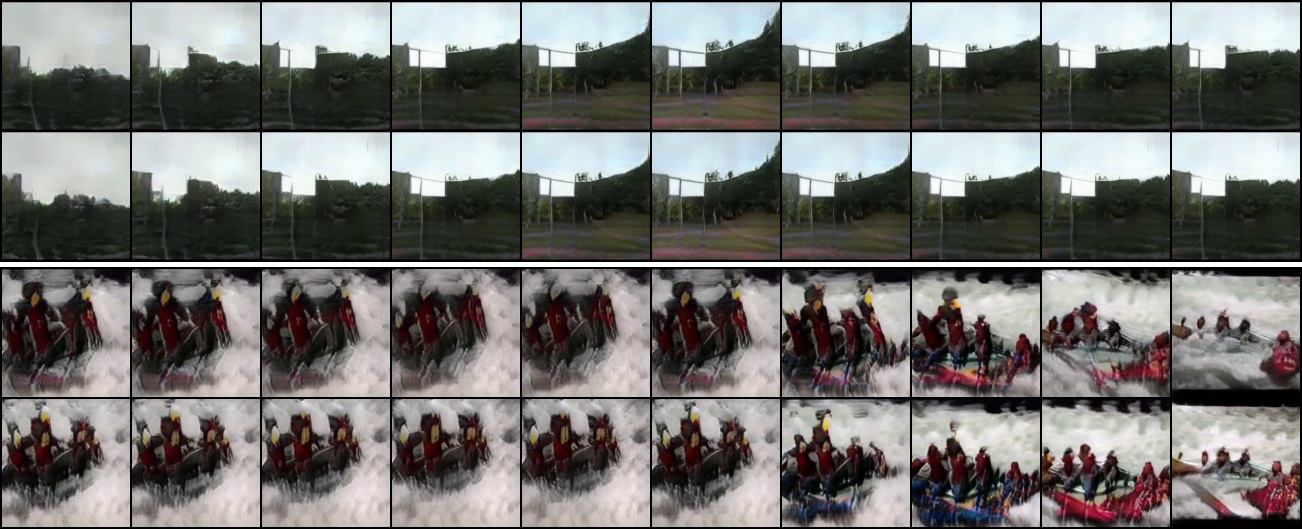}
  \caption{TSB.}
  \label{fig:tmm-intra2}
\end{subfigure}
\caption{Example of intra-class interpolation on UCF-101. The vertical axis represents time, the horizontal axis represents different modes of the class. We sample two latent codes which are represented by the leftmost and right most samples and linearly interpolate between them to generate intermediate latent code samples.}
\vspace{-.8em}
\label{fig:intra}
\end{figure}
\vspace{-.3em}

Figures \ref{fig:quality}, \ref{fig:mais_normal}, and \ref{fig:weiz_gen} show qualitative samples generated by training TSB to generate samples of size $96 \times 96$ on UCF101, MaisToy, and Weizmann datasets respectfully. In Figure \ref{fig:maistoy}, we used the motion label only variant of MaisToy called $\text{MaisToy}_{\text{Single}}$. In this figure we can appreciate that the generation quality is good for all shapes except the triangle. In Figure \ref{fig:maistoy} we showed that TSB had problems generating the triangle shape as well. This might be because of having two different types of triangles in the dataset, filled triangles and empty triangles.

\subsection{Memorization test}
\begin{figure}[h]
\centering
\begin{subfigure}{.95\textwidth}
  \centering
  \includegraphics[width=.95\linewidth]{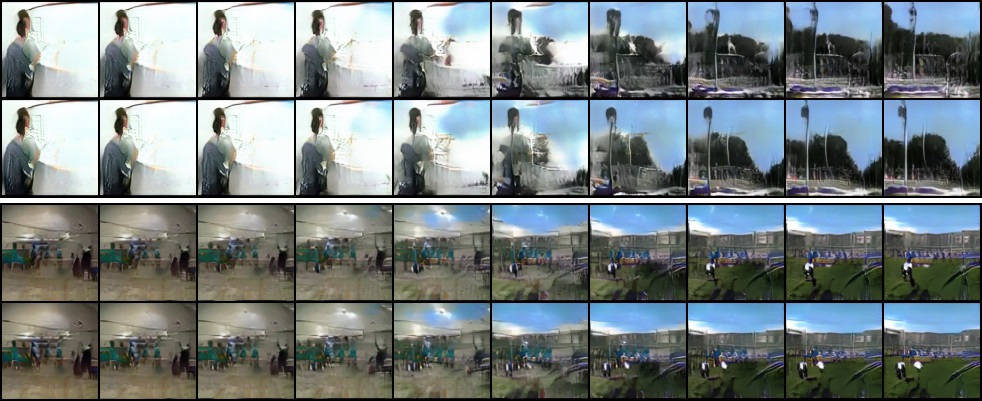}
  \caption{\methodshort.}
  \label{fig:tmm-inter1}
\end{subfigure}%
\vfill
\begin{subfigure}{.95\textwidth}
  \centering
  \includegraphics[width=.95\linewidth]{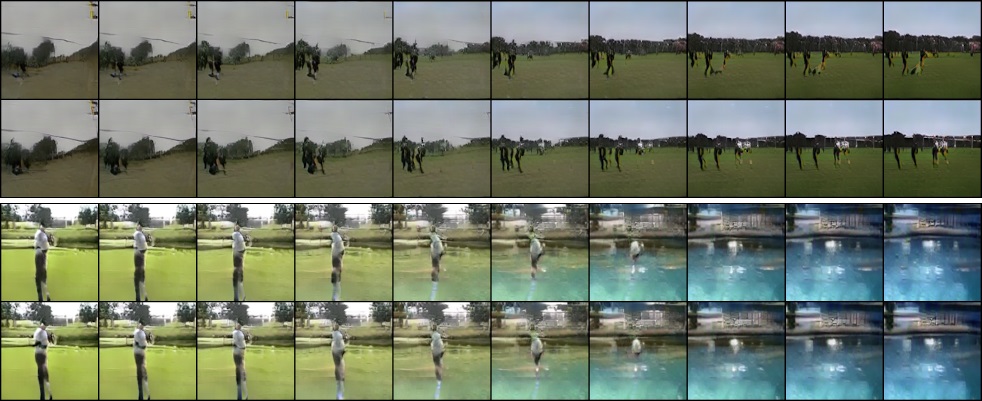}
  \caption{TSB.}
  \label{fig:tmm-inter2}
\end{subfigure}

\caption{Interpolation Performance showing smooth transitions between classes on UCF-101. Each column is a sequence.(a) The top figure is interpolating between classes writing on board (left) and pole vault (right), while the bottom one is interpolating volleyball spiking (left) and frisbee catch (right). (b) The top figure is interpolating between classes basketball (left) and frisbee catch (right), while the bottom one is interpolating between golf swing (left) and diving (right).}
\vspace{-.8em}
\label{fig:inter}
\end{figure}

\begin{figure}[h!]
\centering
\begin{subfigure}{0.95\textwidth}
  \centering
  \includegraphics[width=0.95\linewidth]{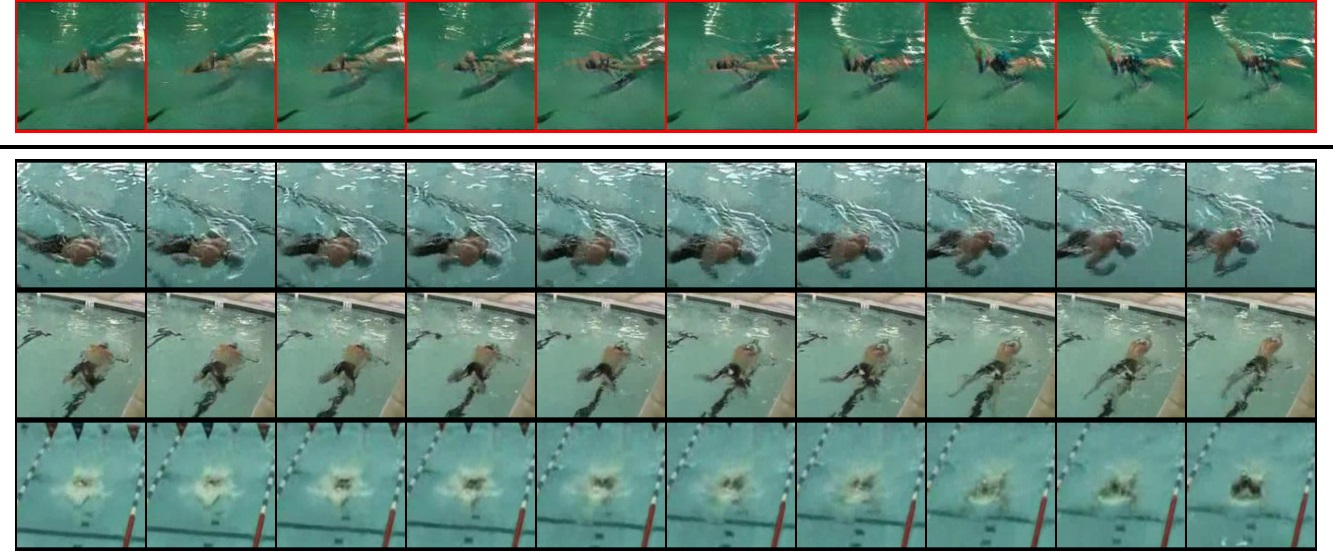}
  \caption{Front Crawl.}
  \label{fig:knn1}
\end{subfigure}%
\vfill
\begin{subfigure}{0.95\textwidth}
  \centering
  \includegraphics[width=0.95\linewidth]{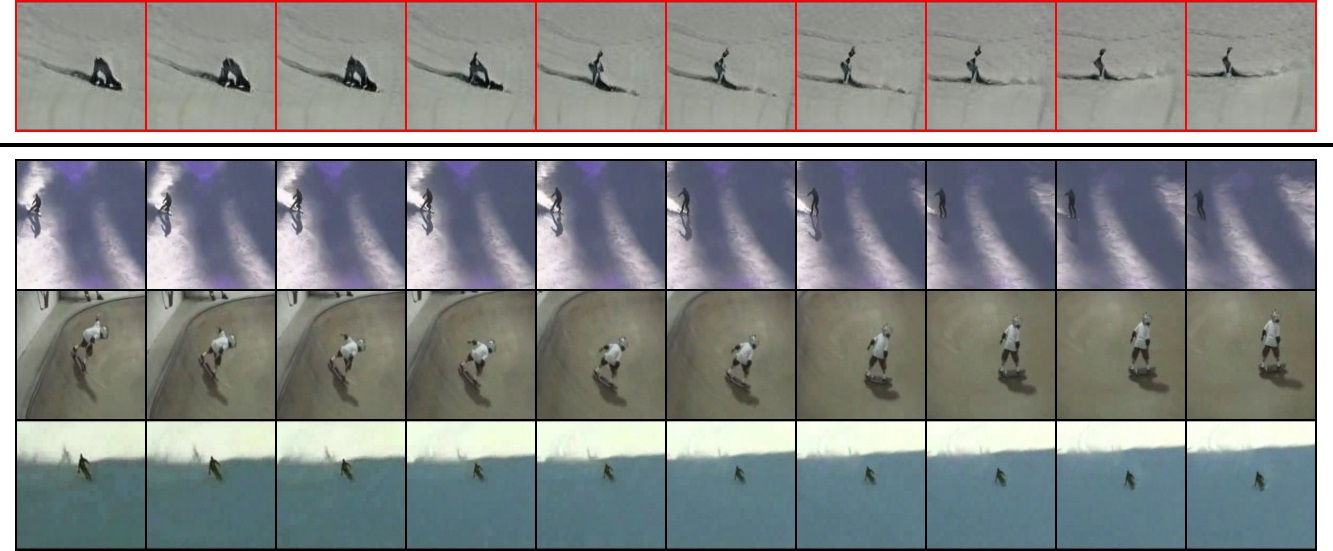}
  \caption{Skiing.}
  \label{fig:knn2}
\end{subfigure}
\vspace{-.4em}
\caption{ Examples of retrieval of top-3 nearest neighbors (black) of TSB generated samples (red). We can see that although the generated samples look similar to their respective 3-NNs, they are still quite visually distinct. This implies the model isn't just memorizing the real data.}
\label{fig:knn}
\end{figure}

\vspace{-1mm}
There is no quantitative measure of memorization in generative models, thus we check this via \emph{intra-class interpolation}, \emph{class interpolation} and k-NN retrieval. In \emph{intra-class interpolation} we linearly interpolate between to two different latent codes $Z_{ABC}$ while keeping the label fixed, as shown in Figure~\ref{fig:intra}. In the Figure~\ref{fig:inter}, we explore \emph{class interpolation} by linearly interpolating between label embeddings, while keeping $Z_{ABC}$ fixed. Figures \ref{fig:intra} and \ref{fig:inter} show smooth transition between modes and classes. If a model would suffer from memorization, we would expect the interpolation to abruptly jump from mode to mode in \emph{intra-class interpolation} and from label to label in \emph{class interpolation}. Samples from the retrieval experiment (Figure \ref{fig:knn}), show that generated samples are noticeable dissimilar to real samples, this suggests that the model does not suffer from memorization. The k-NN experiment was done using the last layer of the ECO \cite{eco} architecture.
\vspace{-1mm}
\section{Conclusion}
\vspace{-2mm}
We presented a TSB architecture for video generation that enforces temporal consistency into a 2D video generator network. We show that TSB design improves the quality of generated videos in comparison to the BigGAN baseline. To validate effectiveness of our method, we conduct experiments on four different datasets including our new dataset MaisToy. Our new dataset enables us to analyze the generalization power of model and also understand which semantics are easier for model to learn.
As a supplement to the well established IS score, we proposed the generalization based S3 score, which is intended to be sensitive to intra-class variation. 
Based on this metric our method also achieves the best performance. These quantitative results are further supported by our qualitative.
\vspace{-2mm}
\section{Acknowledgements}
\vspace{-2mm}
We acknowledge funding by the German Research Foundation (DFG) via the
grant BR 3815/10-1 and the gift of a compute server by Facebook.

\vspace{-2mm}

\FloatBarrier

{\small
\bibliographystyle{ieee_fullname}
\bibliography{egbib}
}

\clearpage

\appendix

\section{Additional Experiments}

\subsection{Generalization test}
In order to evaluate the generalization of model, we held out the following label combinations when training on $\text{MaisToy}_\text{Multi}$ dataset:
\begin{itemize}
    \item Square, red, right
    \item Square, red, left
    \item Square, red, upwards
    \item Square, red, downwards
    \item Triangle, blue, right
    \item Triangle, blue, left
    \item Triangle, blue, upwards
    \item Triangle, blue, downwards
    \item Letter M, yellow, right
    \item Letter M, green, right
    \item Letter M, blue, right
    \item Letter M, red, right
\end{itemize}

\begin{figure}[h]
\centering
\begin{subfigure}{.89\textwidth}
  \centering
  \includegraphics[width=.99\linewidth]{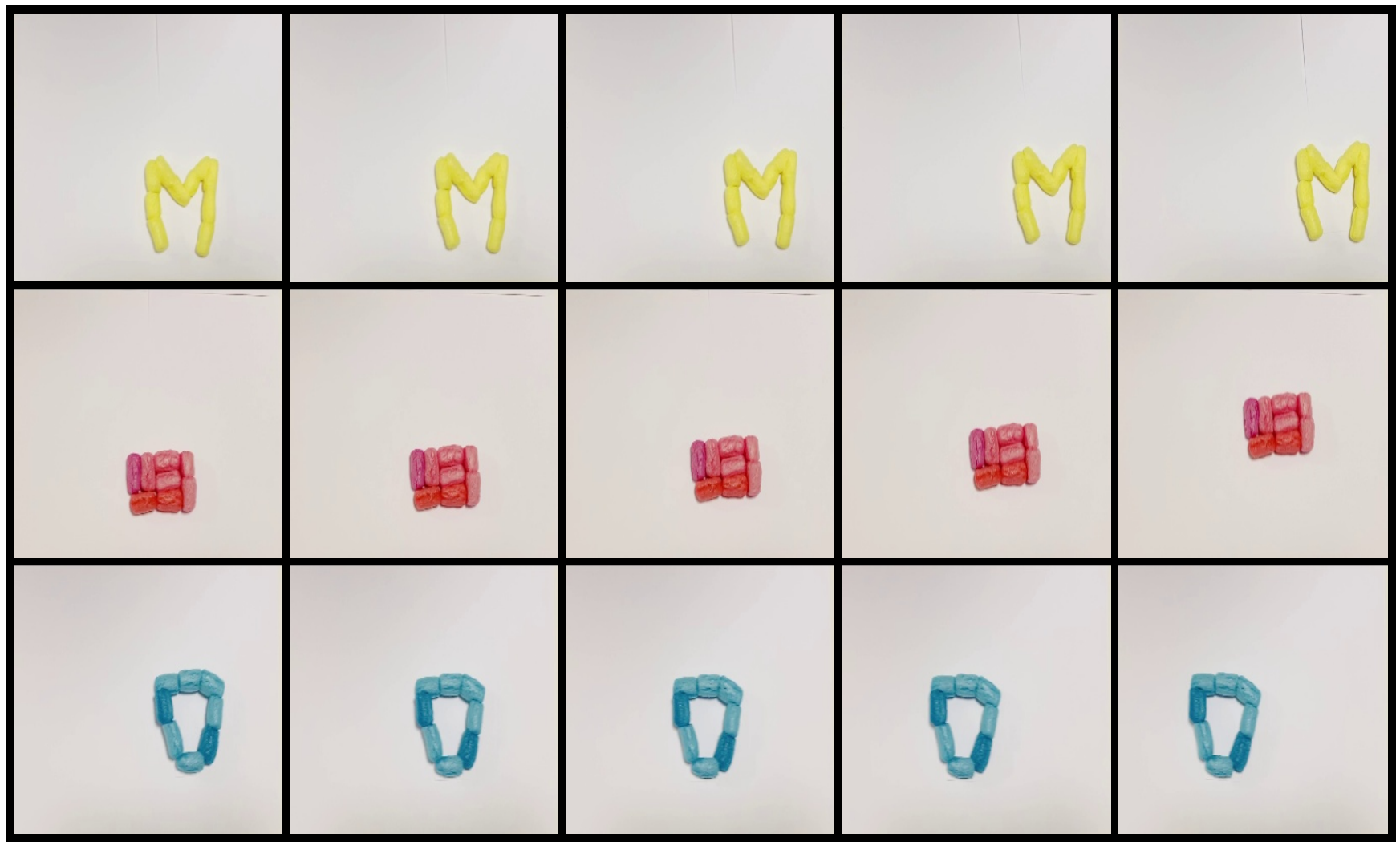}
  \caption{Real sequence.}
  \label{fig:mais1}
\end{subfigure}%
\vfill
\begin{subfigure}{.89\textwidth}
  \centering
  \includegraphics[width=.99\linewidth]{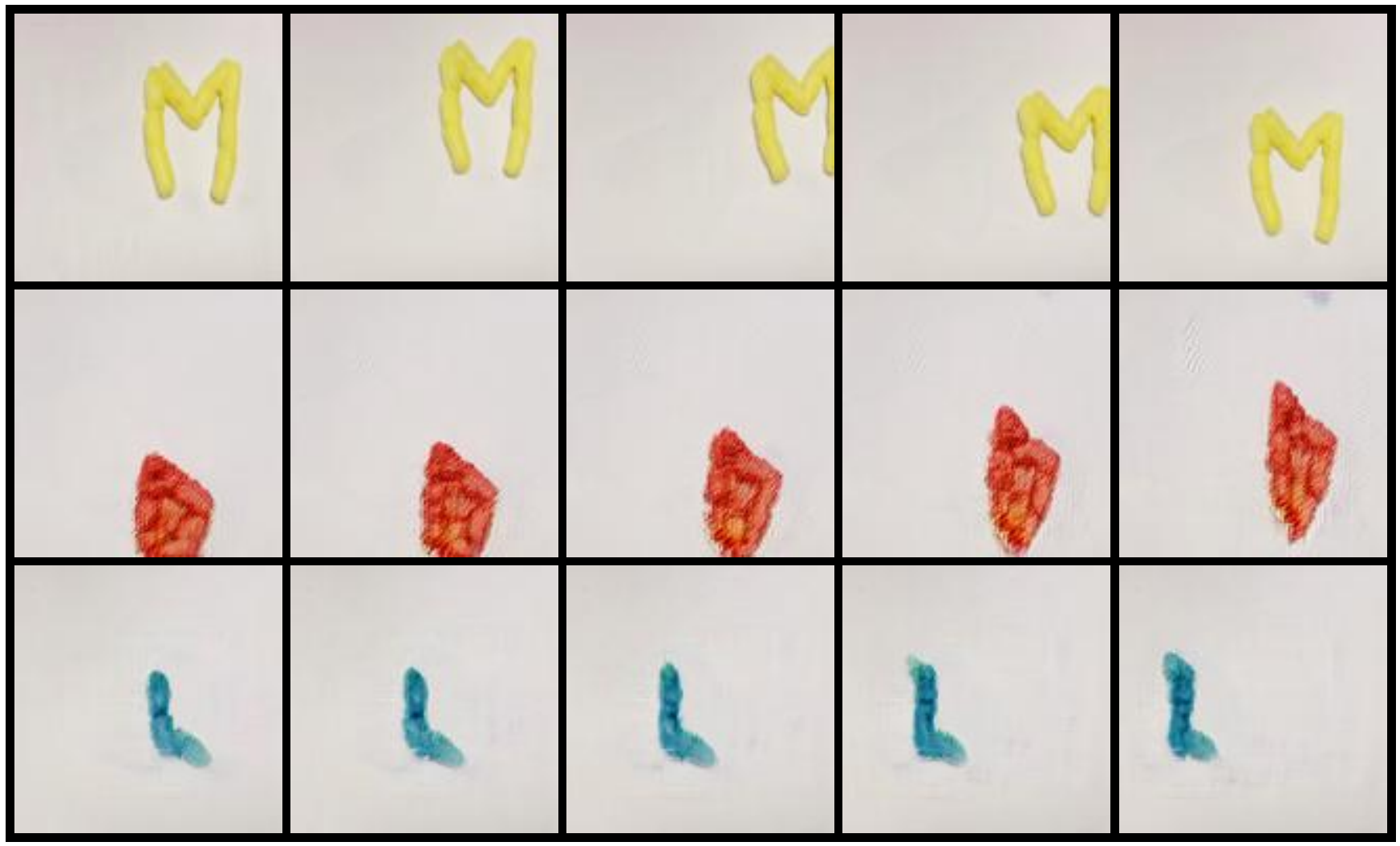}
  \caption{Generated sequence.}
  \label{fig:mais2}
\end{subfigure}
\caption{(a) Samples from held combinations (from top to bottom): Letter M, yellow, right; Square, red, upwards; Triangle, blue, left. (b) Generated sequences for the same combinations presented in (a).}
\vspace{-.8em}
\label{fig:mais}
\end{figure}

Then, we used the network to generate all the unseen combinations above to corroborate if it is actually learning the meaning of each label. Qualitative results in Figure \ref{fig:mais} show that the network is, for the most part, able to generalize to unseen combinations. However, it appears that the triangle shape is difficult for the network.

\subsection{S3}
To calculate S3 we train a classifier on real samples and on fake samples as explained on Section 5. We calculated the S3 metric, for UCF-101 \cite{ucf101}, based on two different classifiers, a TSN  \cite{tsnet} action recognition network and a 3D ResNet-18 \cite{resnet18}. The TSN was trained on a batch size of 14 and an initial learning rate of 0.01. We trained the 3D ResNet-18 using a batch size of 32, an initial learning rate of 0.001 and a dropout probability of 0.6. To get the S3 score for Jester \cite{jester} we decided to use the ECO \cite{eco} action recognition network. We trained it using a batch size of 14, an initial learning rate if 0.001 and a dropout probability of 0.6. All networks were trained on 16 frames and their respective learning rates were scheduled to drop by an order of magnitude after failing to beat the best recorded test accuracy for 4 straight epochs. 

We calculated the S3 with two different architectures for the UCF-101 dataset to provide a reference for the comparisons in the future works. Table~\ref{tab:tsn_vs_3dresnet} shows that the change of architecture does alter the relative performances of ReS and SeR to ReR significantly enough to produce important changes in the score. Therefore, S3 scores obtained from different classification architectures does not provide a fair comparison.

\begin{table}[t]
		\begin{adjustbox}{width=\linewidth,center}
			\begin{tabular}{clcccc}
				%
	\multirow{2}{*}{Classifier Architecture} &\multirow{2}{*}{Method} & Train on: Synth. & \multicolumn{2}{c}{Real}   & \multirow{2}{*}{\textbf{S3}}\\ 
				\cline{3-5}
	      &           & eval. on: Real    & Synth.             & Real &  \\ 
				\hline
	\multirow{2}{*}{{\textbf{TSN}}} 
				& \methodshort     & 45.5 & 46.8 & 85.9 & 0.39\\
				& \textbf{TSB}& 48.55  & 54.91  & 85.9 & 0.45\\
				\hline 
    \multirow{2}{*}{{\textbf{3D ResNet18}}} 
				& \methodshort     & 36.63 & 28.83 & 76.82 & 0.29\\
				& \textbf{TSB} & 44.36  & 29.61  & 76.82 & 0.36\\
				\hline
				
			\end{tabular}
		\end{adjustbox}	\caption{UCF-101 results of S3 on two different architectures.}
		\label{tab:tsn_vs_3dresnet}
		\vspace{-\topsep}
\end{table}

\begin{table}[t]
    \begin{adjustbox}{width=0.8\linewidth,center}
    \begin{tabular}{c c c}
    \hline
          Method & Dataset & FID   \\
          \hline\hline
         \textbf{\methodshort} & UCF-101 & \textbf{3108.77 $\pm$ 0.04} \\
         \hline
         \multirow{2}{*}{\textbf{TSB}}
         & UCF-101 & 3110.29 $\pm$ 0.10 \\
         & Jester & 841.08 $\pm$ 0.005 \\
         \hline
    \end{tabular}
    \end{adjustbox}
    \caption{FID scores on Jester and UCF-101.}
    \label{tab:fid}
\end{table}

\subsection{FID}

FID \cite{fid_score} calculations were done using the features from the second-to-last layer of a TSN  pretrained on Imagenet \cite{imagenet} and finetuned on the respective dataset it is going to be tested on. The network was trained as explained above. We calculated FID using 4000 samples, we repeated the process 5 times to get the standard deviation. Table \ref{tab:fid} seems to suggest \methodshort \: is better than TSB, but this could be due to the fact that FID cannot separate image quality from diversity. If we take into account IS and S3 we can deduce that although FID points to \methodshort \: being better than TSB this is most likely due to better sample diversity, not sample quality.

\subsection{Motion Constraint}

\begin{figure}[h!]
\begin{center}
   \includegraphics[width=0.95\linewidth]{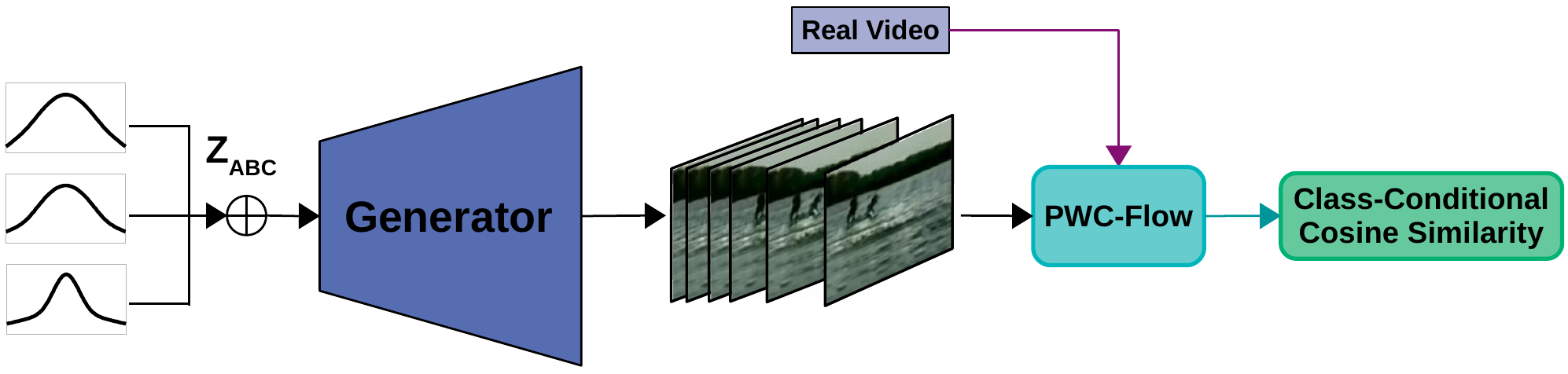}
\end{center}
   \caption{Illustration of the motion constraint calculation. }
\label{fig:cosine}
\end{figure}

Prior work has used optical flow to generate videos by warping the images \cite{wflow1, wflow2} or using it as a prior to generate spatial features \cite{pflow1, tflow}. We rather introduce a intra-class constraint on similarity between optical flows produced by real videos and by generated videos. An illustration is provided in Figure \ref{fig:cosine}.

To estimate this constraint, first we generate synthetic samples and sample real videos from the dataset. Optical flow is calculated using the PWC Flow network \cite{pwc}. Moreover, we calculate the cosine similarity between flows resulted from real videos and flows from generated videos. We only do this for pairs of real and synthetic videos with matching labels:
\vspace{-.2em}
\begin{equation}
    \begin{split}
    L_{M} = \frac{1}{C}\: \Sigma^{B}_{i} \Sigma^{B}_{j} \: \text{Sim}(f_{r_{j}}, f_{f_{i}}) = \\
    \begin{cases}
        \frac{f_{r_{j}} \cdot f_{f_{i}}}{||f_{r_{j}}|| \cdot ||f_{f_{i}}||} & \text  {if $y_{f_{i}} = y_{r_{i}}$} \\
        \varnothing       & \text{otherwise}
    \end{cases}
    \end{split}
    \label{eq:cosine}
\end{equation}
where $f_r$ and $f_f$ stand for real and generated flows, respectively, $B$ is batch size and $C$ is the number of matching real and generated flow pairs. This similarity measure enforces the similarity of motion between samples from the same class. Finally, we add the constraint only to the generator loss:
\vspace{-.2em}
\begin{equation}
    L_{G_{M}} =  L_{G} + (\alpha \cdot (1 -  L_{M}))
    \label{eq:cosine2}
\end{equation}
where $\alpha$ is a hyperparameter that controls the importance of the motion constraint $L_M$.

This architecture was dubbed \methodshort-MC, although it did score better than the baseline with an S3 score of 0.73 on the Weizmann dataset, it fell short of \methodshort-VAR and TSB. Among some other disadvantages of this motion constraint is the fact that it makes training significantly slower and unstable.

\subsection{Ablation studies}

\begin{figure}[h]
\begin{center}
   \includegraphics[width=\linewidth]{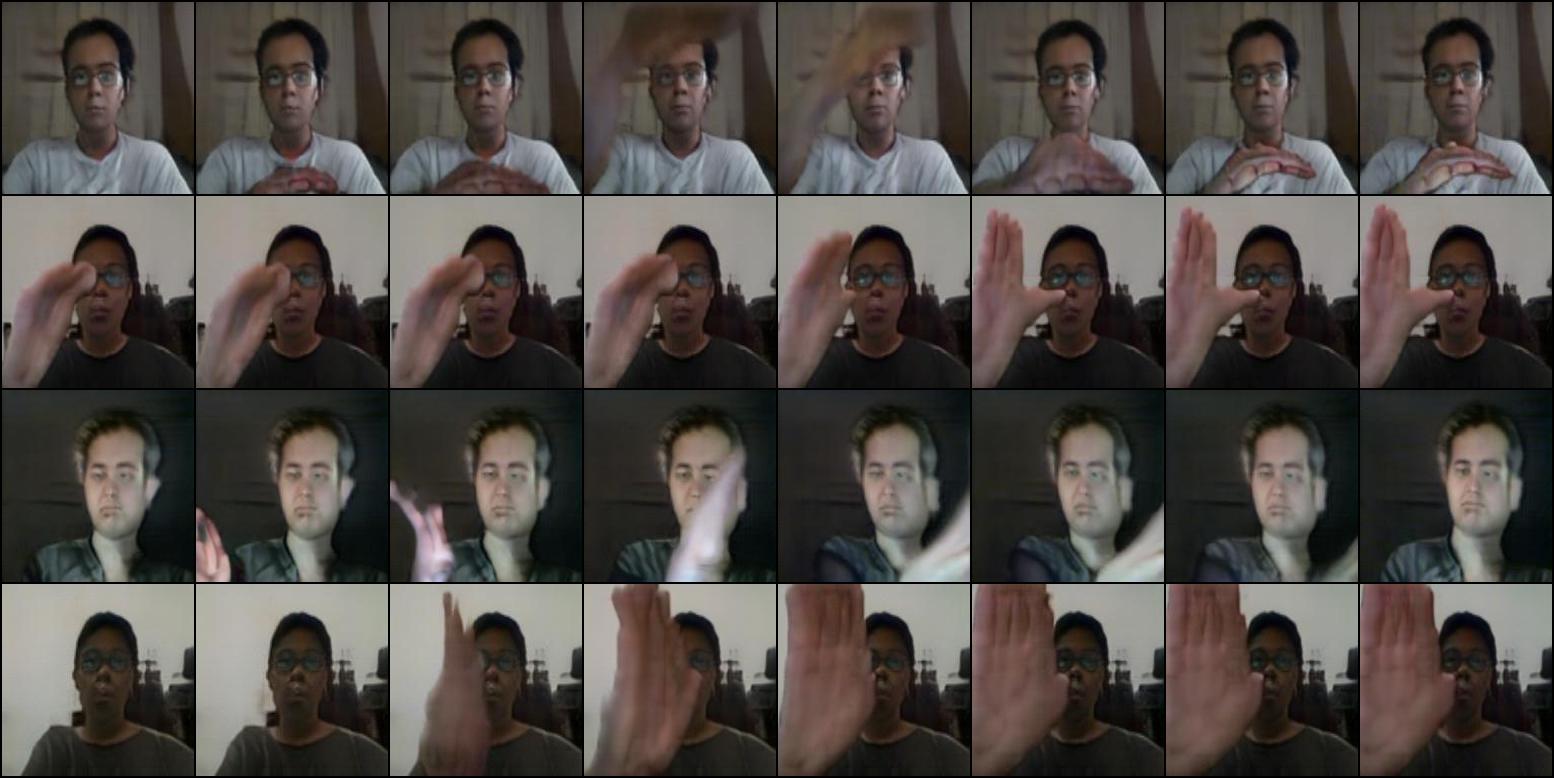}
\end{center}
   \caption{Generated samples of TSB trained to produce $192 \times 192$ samples of the Jester dataset.}
   \vspace{-1em}
\label{fig:jest_192}
\end{figure}

Our TSB trained on Jester did not record a good performance on the S3 measure, hence we need a qualitative evaluation to look for a possible reason why this was the case. Figure \ref{fig:jest_192} shows an acceptable level of quality in both spatial and motion feature generation. However, TSB still was not able to produce realistic enough samples in fine structures of hands and faces as a real person would have.



\begin{figure}[h]
\begin{center}
   \includegraphics[width=\linewidth]{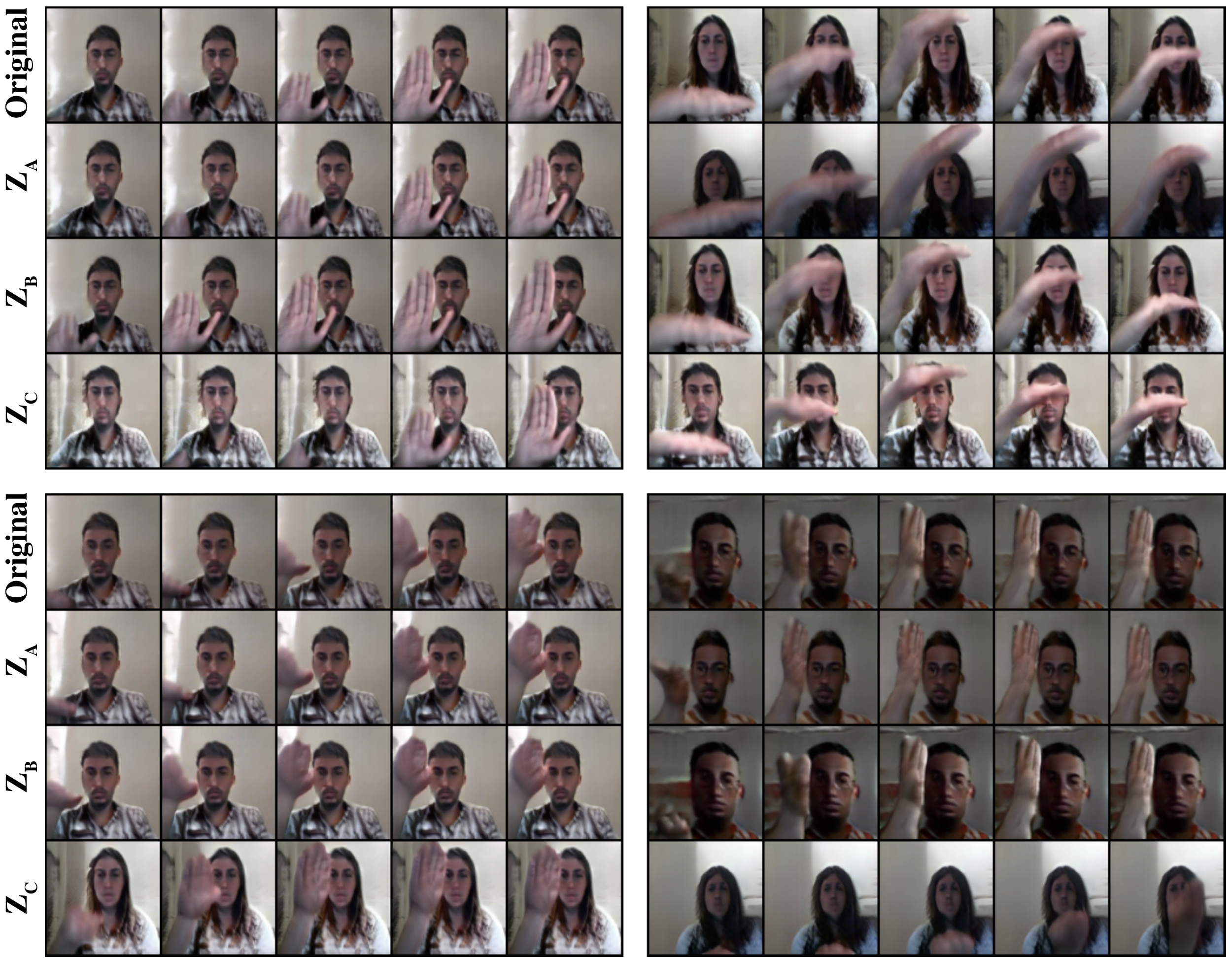}
\end{center}
   \caption{Latent variable experiment. We freeze two out the three subspaces and re-sample the remaining one to produce a new sample. We compare each sample to the original to see what meaning is the network assigning to that specific subspace.}
   \vspace{-1em}
\label{fig:noise_fig}
\end{figure}

\begin{figure*}
\begin{center}
\includegraphics[width=0.95\linewidth]{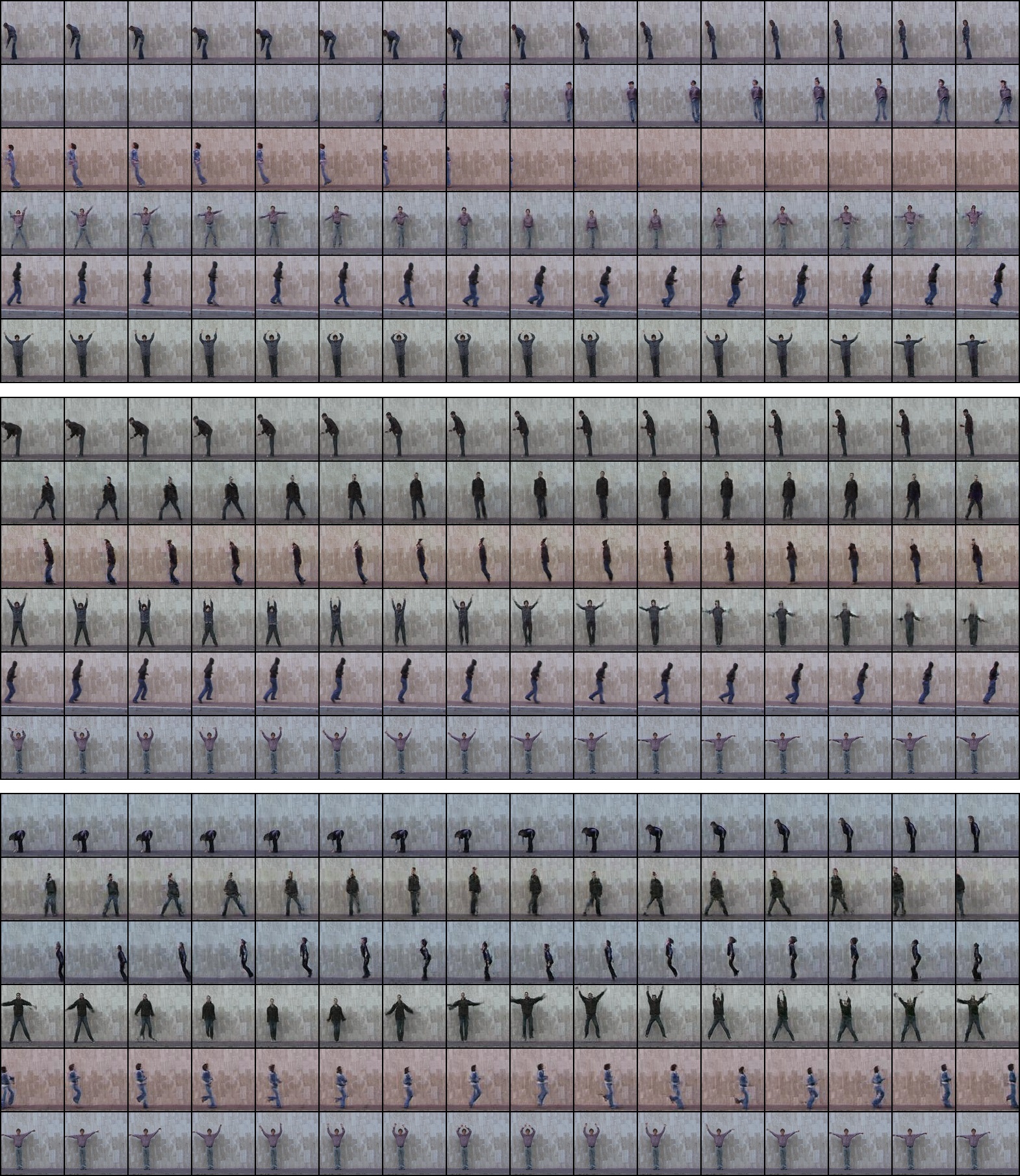}
\end{center}
\vspace{-.4em}
   \caption{Generated samples (from top to bottom) of \methodshort \:, \methodshort-MC, \methodshort-VAR trained on Weizmann.}
   \vspace{-.8em}
\label{fig:quality_weiz}
\end{figure*}

\paragraph{Impact of latent codes.}
We wanted to know if in fact using a multi-variate model for the latent codes had any effect on what the network learned. Specifically we wanted to see if assigning a different variance to each subspace had any effect on the features the network learned to map to each one of the subspaces. To test this we froze two out of the three subspaces and re-sampled the remaining one to produce a new sample. Every subspace will get a turn at being re-sampled. Figure \ref{fig:noise_fig} shows some examples of this experiment compared to a sample produced by the originally sampled latent vector. The samples show that the network learns to assign $Z_C$ features that result in bigger changes in the overall visual features, like gender. We can observe as well that $Z_B$ appears to be in charge more of motion features, without affecting features such as location or person identity as much. It appears that $Z_A$ is in charge of more infrequent features like location or small changes in appearance. This experiment points towards the variance assigned to a subspace being directly related the types of features it represents.
\paragraph{Ablation study on different model designs.}
Figure \ref{fig:quality_weiz} shows the improvement between models in different classes of the Weizmann \cite{weiz} dataset. 

\section{Architectural Details}

We adopted most of BigGAN's \cite{biggan} architectural choices in $G_{\text{Image}}$, with the exception that we moved the self-attention module down one level of abstraction to save video memory. $D_{\text{Image}}$ follows exactly the discriminator guidelines set in BigGAN, while $D_{\text{Video}}$ adopted the exact architecture used in MoCoGAN \cite{moco}, but extended for class conditional hinge loss per \cite{cgans}. To describe the width of all networks we use the product of a layer-wise constant $c$ and a per-layer constant $a$. In all experiments $a$ was set to 96. We chose $c$ to be $[16, 8, 4, 2, 1]$ for $G_{\text{Image}}$, $[1, 2, 4, 8, 16, 16]$ for $D_{\text{Image}}$ and $[1, 2, 4, 8]$ on $D_{\text{Video}}$.

At the input of $G_{\text{Image}}$ we have a fully connected layer which applies an affine transformation to $Z_F$ to transform it from $[T, d+120]$ to $[T,w \cdot h \cdot 16 \cdot a]$. When generating $96 \times 96$ sized samples we set $w$ and $h$ to 3 and when we generated samples of size $128 \times 128$ they were both set to 4.

The sequence generator is composed of a fully connected layer $FC$ and a GRU cell. $FC$ has a size of $d$ and the GRU cell has a size of 2048.

\end{document}

%% file: table1.tex
\begin{table*}
           \centering
           \captionsetup[subtable]{position = below}
           \begin{subtable}{0.40\linewidth}
               \centering
               \begin{adjustbox}{width=\linewidth,center}
     \begin{tabular}{l l l}
            \hline
        Dataset & Method & IS $(\uparrow)$\\ [0.15ex]
        \hline\hline
        \multirow{10}{4em}{UCF-101}& VGAN \cite{vgan}   & 8.18 $\pm$ 0.09\\ [0.15ex]
        
        & TGAN~\cite{tgan}  & 11.85 $\pm$ 0.07\\[0.15ex]
        
        & MDP~\cite{markov}  & 11.86 $\pm$ 0.11\\[0.15ex]
        
        & MoCoGAN~\cite{moco}   & 12.42 $\pm$ 0.03 \\[0.15ex]
        
        & Prog. VGAN~\cite{progrevgan}   & 14.56 $\pm$ 0.05 \\[0.15ex] 

        & TGANv2~\cite{tganv2}   & 24.34 $\pm$ 0.35 \\[0.15ex]
        & DVD-GAN~\cite{dvdgan}  & 32.97 $\pm$ 1.7 \\[0.15ex]
        & \textbf{\methodshort}  & 31.91 $\pm$ $0.14$\\[0.15ex]
        
        & \textbf{TSB}  & \textbf{42.79 $\pm$ 0.63} \\[0.15ex]
        
        \hline
        Jester & \textbf{TSB}  & \textbf{11.90 $\pm$ 0.12} \\
        \hline
    \end{tabular}
    \end{adjustbox}
    \caption{Comparison of Inception Score between models on UCF-101. }
    \label{tab:isscore}
           \end{subtable}%
           \hspace*{1em}
           \begin{subtable}{0.57\linewidth}
               \centering
               \begin{adjustbox}{width=\linewidth,center}
			\begin{tabular}{clcccc}
				\hline
	\multirow{2}{*}{Dataset} &\multirow{2}{*}{Method} & Train on: Synth. & \multicolumn{2}{c}{Real}   & \multirow{2}{*}{\textbf{S3}}\\ 
				\cline{3-5}
	      &           & eval. on: Real    & Synth.             & Real &  \\ 
				\hline\hline
	\multirow{4}{*}{\textbf{Weizmann}} &
				MoCoGAN           & 61.11 & 32.83  & 95.83 & 0.37\\
				&\methodshort     & 80.31 & 62.53  & 95.83 & 0.67\\
				&\methodshort-VAR  & \textbf{88.30} & 64.29  & 95.83 & 0.75 \\
				&TSB & 88.10  & \textbf{71.43} & 95.83 & \textbf{0.79}\\
				\hline
	$\textbf{MaisToy}_{\text{Multi}}$
                & TSB & \textbf{82.74} & \textbf{45.11} & 67.70  & \textbf{0.99}\\
                \hline
    $\textbf{MaisToy}_{\text{Single}}$
                & TSB & \textbf{45.83} & \textbf{52.98} & 66.07  & \textbf{0.62}\\
                \hline
	\multirow{2}{*}{{\textbf{UCF101}}} 
				& \methodshort     & 45.5 & 46.8 & 85.9 & 0.39\\
				& TSB & \textbf{48.55}  & \textbf{54.91}  & 85.9 & \textbf{0.45}\\
				\hline 
    \textbf{Jester} 
                & TSB & \textbf{15.85} & \textbf{13.86} & 91.4  & \textbf{0.07}\\
				\hline
			\end{tabular}
		\end{adjustbox}
		\vspace{0.8em}
		\caption{Evaluation S3 measure based on action classification generalization between generated and real samples.}
		\label{tab:classification_test}
           \end{subtable}
                      \caption{\textbf{Comparison with previous methods on four different datasets.}}
       \end{table*}

\begin{table}[h!]
		\begin{adjustbox}{width=\linewidth,center}
			\begin{tabular}{cccccccc}
				\hline
	\multirow{2}{*}{Low Quality} &\multirow{2}{*}{Collapsed} &\multirow{2}{*}{Normal} & Train on: Synth. & \multicolumn{2}{c}{Real}   & \multirow{2}{*}{\textbf{S3}} & \multirow{2}{*}{\textbf{IS $(\uparrow)$}}\\ 
				\cline{4-6}
	      &           & &  eval. on: Real    & Synth.             & Real &  &\\ 
				\hline\hline
	 \xmark     & & & 41.4 & 50.06 & 85.9 & 0.36 & 34.01 $\pm$ 0.22\\
				  & \xmark & & 28.3  & 59.3  & 85.9 & 0.27 & 41.04 $\pm$ 0.82\\
				 & & \xmark &\textbf{48.55}  & \textbf{54.91}  & 85.9 & \textbf{0.45} & \textbf{42.79 $\pm$ 0.63}\\
				\hline 
			\end{tabular}
		\end{adjustbox}	\caption{Comparison between TSB's UCF-101 S3 and IS results at different levels of model quality.}
		\label{tab:s3_test}
		\vspace{-\topsep}
\end{table}